\documentclass[conference,a4paper]{IEEEtran}
\IEEEoverridecommandlockouts

\usepackage[hidelinks]{hyperref}
\usepackage[cmex10]{amsmath}
\usepackage{amssymb,amsfonts}
\usepackage{dblfloatfix}

\usepackage[ruled,vlined]{algorithm2e}
\usepackage{graphicx}
\usepackage{booktabs}
\usepackage{siunitx}

\usepackage[numbers,compress]{natbib}

\usepackage{texnames}
\usepackage{bm,bbm}
\usepackage{orcidlink}

\usepackage{textcomp}
\usepackage{xcolor}
\usepackage{url}
\usepackage[caption=false,font=normalsize,labelfont=sf,textfont=sf]{subfig}
\usepackage{soul}
\soulregister\cite7
\soulregister\citet7
\soulregister\citep7
\soulregister\ref7
\soulregister\pageref7

\usepackage{multirow}
\usepackage{svg}
\usepackage{balance}
\usepackage{comment}

\begin{document}

\title{\uppercase{Histogram-based Parameter-efficient Tuning for Passive and Active Sonar Classification}
\thanks{DISTRIBUTION STATEMENT A. Approved for public release. Distribution is unlimited.
This material is based upon work supported by the Under Secretary of War for Research and Engineering under Air Force Contract No. FA8702-15-D-0001 or FA8702-25-D-B002. Any opinions, findings, conclusions or recommendations expressed in this material are those of the author(s) and do not necessarily reflect the views of the Under Secretary of War for Research and Engineering.
\copyright 2026 Massachusetts Institute of Technology.

Delivered to the U.S. Government with Unlimited Rights, as defined in DFARS Part 252.227-7013 or 7014 (Feb 2014). Notwithstanding any copyright notice, U.S. Government rights in this work are defined by DFARS 252.227-7013 or DFARS 252.227-7014 as detailed above. Use of this work other than as specifically authorized by the U.S. Government may violate any copyrights that exist in this work.}
}

\author{
\IEEEauthorblockN{Amirmohammad Mohammadi$^{1}$,
Davelle Carreiro$^{2}$,
Alexandra Van Dine$^{2}$,
Joshua Peeples$^{1}$\orcidlink{0000-0001-7861-7789}}

\IEEEauthorblockA{\textit{$^{1}$Department of Electrical and Computer Engineering, Texas A\&M University, College Station, TX, USA}}
\IEEEauthorblockA{\textit{$^{2}$Massachusetts Institute of Technology Lincoln Laboratory, Lexington, MA, USA}}
}

\maketitle
\begin{abstract}
Parameter-efficient transfer learning (PETL) methods adapt large artificial neural networks to downstream tasks without fine-tuning the entire model. However, existing additive methods, such as adapters, sometimes struggle to capture distributional shifts in intermediate feature embeddings. We propose a novel histogram-based parameter-efficient tuning (HPT) technique that captures the statistics of the target domain and modulates the embeddings. Experimental results on three downstream passive sonar datasets (ShipsEar, DeepShip, Vessel Type Underwater Acoustic Data (VTUAD)) demonstrate that HPT outperforms conventional adapters. Notably, HPT achieves 91.8\% vs. 89.8\% accuracy on VTUAD. For active sonar imagery (Watertank, Turntable), HPT is competitive with other PETL methods.
Furthermore, HPT yields feature representations closer to those of fully fine-tuned models. Overall, HPT balances parameter savings and provides a distribution-aware alternative to existing adapters and shows a promising direction for transfer learning in resource-constrained environments. The code is publicly available: \url{https://github.com/Advanced-Vision-and-Learning-Lab/HLAST_DeepShip_ParameterEfficient}.
\end{abstract}


\section{Introduction}
\label{introduction}

Underwater acoustic recognition has various applications in marine environments such as search and rescue, seabed mapping, and shipping traffic monitoring \cite{testolin2020detecting, beckler2022multilabel}. Developing reliable recognition systems in this domain remains challenging due to the scarcity of labeled underwater data and the high variability of acoustic conditions \cite{xie2024advancing}. Transfer learning from pre-trained models has shown performance improvements compared to training from scratch, especially in scenarios with limited labeled data \cite{pegeot2023comprehensive}. Yet, fine-tuning all parameters of large-scale models for each downstream task is resource-intensive, as a new set of weights must be tuned and stored for each target task, making this impractical for edge-device applications \cite{evci2022head2toe}. To address these issues, parameter-efficient transfer learning (PETL) methods \cite{han2024parameter} have been developed, where the pre-trained model’s weights are kept frozen while only a small set of parameters are optimized to adapt to the target task.

Many PETL techniques are based on methods such as inserting adapters \cite{houlsby2019parameter}, adding prompt tokens \cite{lester2021power}, or applying low-rank decompositions \cite{hu2021lora}. Although these approaches have demonstrated strong empirical performance, some tasks may benefit from more distribution-aware adaptations \cite{lian2022scaling}. In particular, acoustic data can face distribution mismatches due to diverse recording conditions and environmental factors \cite{mesaros2018multi}. Mitigating these mismatches through a distribution-aware adaptation is one potential approach to improve generalizability. In this work, a histogram-based parameter-efficient tuning (HPT) method is proposed that learns to represent intermediate embeddings as feature distributions.

\section{Related work}
\label{relatedwork}

Following \cite{han2024parameter}, PETL methods can be categorized into four groups: additive, selective, reparameterized, and hybrid. Additive methods such as adapters \cite{houlsby2019parameter,he2021towards} introduce small trainable modules into the pre-trained model. Soft prompts like prefix-tuning \cite{li2021prefix} and prompt-tuning \cite{lester2021power} add learnable vectors to the input sequence. Selective methods only update a subset of existing parameters, such as BitFit \cite{zaken2021bitfit}, which tunes only a set of bias terms. Reparametrized methods (e.g., low-rank adaptation (LoRA) \cite{hu2021lora}) learn compact updates that can be merged back into the original weights at inference. Scaling and shifting features (SSF) \cite{lian2022scaling} inserts additional layers after every major operation to apply linear transformations to the features. Finally, hybrid methods unify multiple strategies, acknowledging that different tasks may require different forms of adaptation. For example, UniPELT \cite{mao2021unipelt} integrates LoRA, prefix-tuning, and adapters, using a gating mechanism to modulate their relative contributions. PETL research remains less explored in underwater acoustic domain.

SSF \cite{lian2022scaling} assumes that the upstream datasets and downstream datasets have different data distributions \cite{sun2016return} thereby justifying a reparamterized linear transformation to alleviate this mismatch. Similarly, this manuscript proposes an additive non-linear transformation to capture feature statistics. This work seeks to advance beyond conventional adapters by incorporating distribution-aware mechanisms using histogram layers \cite{peeples2021histogram} on passive and active sonar in underwater monitoring \cite{tian2021deep}. Using histogram layers for passive sonar has been shown to improve performance of convolutional neural networks \cite{ritu2023histogram}, while acoustic models face performance degradation due to distributional mismatches between training and test conditions \cite{wang2017unsupervised, liu2023advancing}. 
\section{Method}
\label{sec:method}

The original histogram layer \cite{peeples2021histogram} approach was used for 2-dimensional data (i.e., images). The histogram layer herein is adapted for 1-dimensional data (\textit{i.e.}, sequences) and incorporates this method parallel to the multi-head self-attention (MHSA) sublayer of the transformer architecture.
{The MHSA sublayer computes token-to-token dependencies by deriving Query ($\mathbf{Q}$), Key ($\mathbf{K}$), and Value ($\mathbf{V}$) representations from the same normalized input sequence. It then applies the scaled dot-product attention mechanism defined as $\text{Softmax}(\mathbf{Q}\mathbf{K}^T/\sqrt{d_k})\mathbf{V}$, where $d_k$ is the dimension of the key vectors per attention head (i.e., the model dimension $D$ divided by the number of heads). This term serves as a scaling factor to prevent softmax saturation that would cause vanishing gradients.}
Let \(\mathbf{X} \in \mathbb{R}^{N \times D}\) denote the input to the histogram layer, where \(D\) is the feature dimension (e.g., 768) and \(N\) is the sequence length (i.e., number of patches). The histogram layer first applies two successive \(1\times1\) convolutions. The first convolution projects \(\mathbf{X}\) onto \(B\) channels (i.e., bins), with the learnable biases serving as bin centers \(\{\mu_b\}_{b=1}^{B}\). The second, implemented as a grouped convolution, learns bin-specific widths \(\{\gamma_b\}_{b=1}^{B}\). For each feature value \(x\), a radial basis function (RBF) computes its assignment to the \(b\)th bin as shown in \eqref{eqn:binning}:

\begin{equation}
    y_b(x) = \exp\Bigl(-\gamma_b^2\, (x-\mu_b)^2\Bigr),
    \label{eqn:binning}
\end{equation}

Here, \(x\) denotes the per-patch, per-bin value produced by the first convolution (i.e., \(x = \mathbf{x}\mathbf{w}_{b}\)). Responses are then normalized across bins so the contribution of each descriptor is in the range of \([0, 1]\). Specifically, the normalized bin response is defined in \eqref{eq:bin_norm}:

\begin{equation}
    \hat{r}_b(x) = \frac{y_b(x) }
    {\displaystyle \sum_{b'=1}^{B}y_b(x)  + \epsilon}, \quad b=1,\dots,B,
    \label{eq:bin_norm}
\end{equation}

\noindent where \(\epsilon\) is a small constant (\(10^{-6}\)) added for numerical stability to prevent division by zero. The data shape is transformed to $N$ patches that are $B$-dimensional. These normalized responses are then aggregated via adaptive average pooling along the sequence dimension to capture the overall statistical distribution. Equation \eqref{eq:pool_broadcast} denotes the pooling and broadcasting: 

\begin{equation}
    \mathcal{H}(\mathbf{X}) = \operatorname{Broadcast}\!\Bigl\{
    \mathcal{P}\!\Bigl[\{\hat{r}_b(x)\}_{b=1}^{B}\Bigr] \Bigr\}.
    \label{eq:pool_broadcast}
\end{equation}

\begin{figure}[t]
    \centering
    \centerline{\includegraphics[width=0.95\columnwidth]{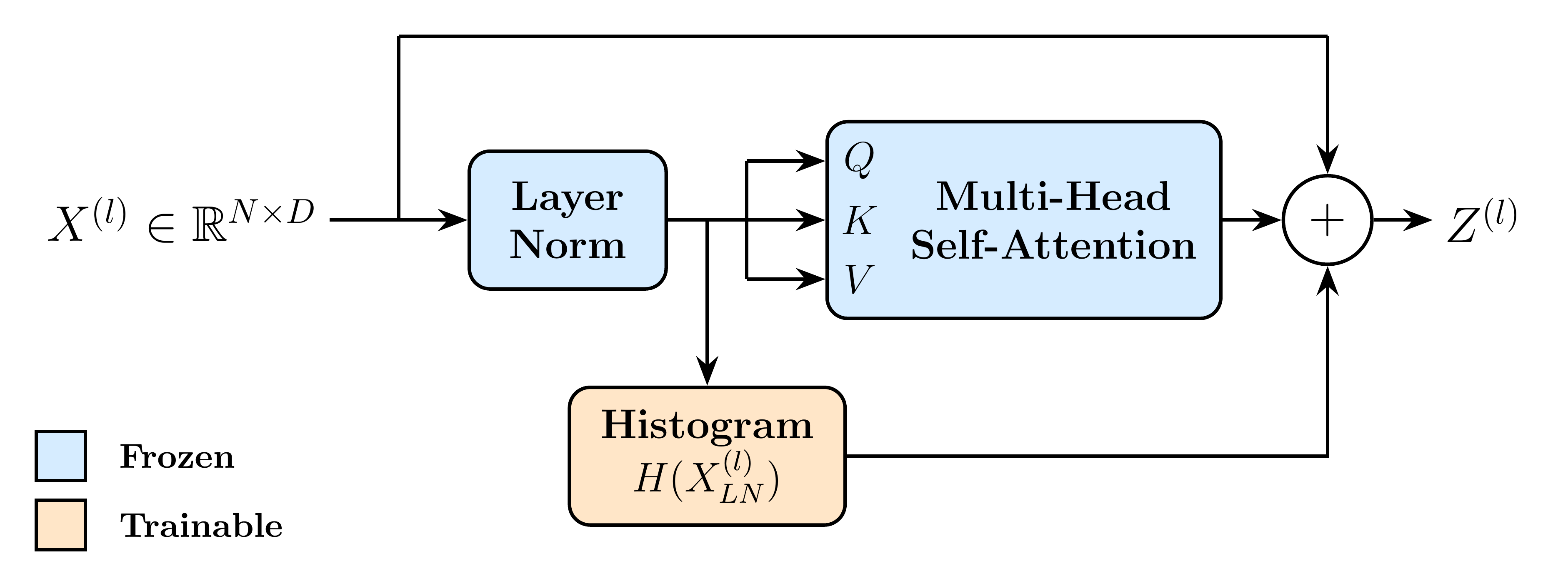}}
    \caption{
        Integration of the histogram layer into the transformer architecture. The histogram layer operates in parallel with the MHSA sublayer.}
    \label{fig:histogram_integration}
\end{figure}

\noindent Here, \(\mathcal{P}(\cdot)\) is the adaptive average pooling operation that reduces the variable-length responses to a fixed-size summary, and \(\operatorname{Broadcast}(\cdot)\) replicates this summary across all patches so that every patch is augmented with the same statistical context. Adaptive average pooling reduces only the sequence axis \(N \rightarrow P\) (with \(P = D/B\)). Flattening \((P,B) \mapsto (PB=D)\) and broadcasting this vector to every patch restores the shape to \(N \times D\), matching the model dimension. Given \(\mathbf{X}_{\text{LN}}\) and the histogram-based representation, the updated output (as shown in Fig. \ref{fig:histogram_integration}) is computed as:

\begin{equation}
\mathbf{Z} = \mathbf{X} + \text{MHSA}(\mathbf{X}_{\text{LN}}) + \mathcal{H}(\mathbf{X}_{\text{LN}}).
\end{equation}

While the described approach applies the histogram layer after layer normalization and incorporates its output alongside the MHSA sublayer’s features, other configurations are possible. For instance, the histogram layer may be applied after the feed-forward network (FFN) sublayer. Histogram features were added alongside MHSA to inform the self-attention process with distributional statistics. In practice, histogram features may be introduced at multiple portions of the network (in parallel and/or residual). 

\section{Experimental Setup}
\label{exp}

An evaluation of the histogram-based adaptation approach against full fine-tuning, linear probing, and adapter methods is presented in this section. The Audio Spectrogram Transformer (AST) \cite{gong2021ast} (built upon a Vision Transformer (ViT) \cite{touvron2021training}) which is pre-trained on ImageNet \cite{deng2009imagenet} and AudioSet \cite{gemmeke2017audio} is used. Experiments were conducted on three passive sonar datasets: ShipsEar \cite{SANTOSDOMINGUEZ201664}, DeepShip \cite{IRFAN2021115270}, and vessel type underwater acoustic data (VTUAD) \cite{Domingos2022}.
For ShipsEar and DeepShip, recordings (89 for ShipsEar; 609 for DeepShip) were resampled to 16 kHz and segmented into non-overlapping 5-second examples, yielding 2,194 segments for ShipsEar and 33,770 segments for DeepShip. The segments are then partitioned by recording into 70\% train, 10\% validation, and 20\% test splits such that all segments from any recording remain in the same partition. VTUAD provides 175,965 32 kHz 1-second examples with predefined splits. Each input signal is transformed into a 128-dimensional log Mel-frequency spectrogram \cite{kong2019weakly} with a window length of 2048 and a hop length of 512.

Also, two active sonar datasets were evaluated: Watertank and Turntable \cite{singh2021marine}. In Watertank, diverse objects (e.g., bottles, cans, tires, valves) were placed on a tank floor while the forward looking sonar (FLS) captured them from multiple viewpoints. In Turntable, the sensor was fixed and objects were rotated to span a full 360° aspect. Each dataset consists of single-channel 96 by 96 pixels with values normalized between [0,1]; Watertank has 11 classes with splits 1838/394/395 (train/val/test), and Turntable has 12 classes with splits 1349/156/645. Following the original paper \cite{singh2021marine}, the training-split mean is subtracted per-sample for normalization and the data augmentation is applied with random shifts up to 10\% and horizontal flips with probability 0.5.

For passive and active sonar experiments, the learning rate is set to \(1e^{-5}\) for full fine-tuning and \(1e^{-3}\) for all other methods. The optimizer is AdamW and the batch size is 64 for ShipsEar and DeepShip, and 256 for VTUAD. The batch size is set to 32 for Watertank and Turntable. The maximum epoch number is set to 200 and early stopping is implemented to stop training if there is no improvement in the validation loss after 20 epochs. The reported accuracies are based on the unseen test split. Mean accuracy (\%) along with standard deviations (subscripted) from three runs of random initialization are reported. The histogram convolutional layers use PyTorch’s default initialization (Kaiming method), while the adapters are zero-initialized. Experiments were completed on an A100 GPU.

\section{Results and Discussion}
Initially, HPT is compared with adapters since both belong to the same additive category. Three adapter configurations are considered, defined by reduction rates of 256, 128, and 64. Three histogram configurations are considered, defined by 4, 8, and 16 bins. Initially, a shared-weights setting where the same adapter or histogram parameters are used across all transformer layers is applied. This range of configurations is chosen to systematically explore the trade-off between accuracy and parameter efficiency, so that both lightweight and more expressive modules are compared. Table \ref{tab:results_shared} presents results under this scenario, showing that the histogram configurations outperform adapters. More parameters leads to a higher accuracy, thus indicating the scalability of the HPT method.

\begin{table}[ht]
\centering
\caption{Mean classification accuracy (\%) with subscripted standard deviations and total trainable parameters under the shared-weights setting. The best accuracies are shown in \textbf{bold}. In the Params column, the left number corresponds to DeepShip, and the right number corresponds to ShipsEar and VTUAD.}
\label{tab:results_shared}
\renewcommand{\arraystretch}{1.2}
\begin{tabular}{l c c c c}
\toprule
Method           & DeepShip        & ShipsEar        & VTUAD           & Params \\
\midrule
Full fine tune      & $70.0_{1.0}$    & $60.0_{1.4}$    & $\mathbf{96.3_{0.1}}$  & 86.9M/86.9M \\
Linear probe        & $66.2_{0.8}$    & $56.4_{2.5}$    & $79.9_{0.2}$           & 4.9K/5.6K \\
Adapter (256)       & $67.1_{0.7}$    & $56.5_{1.7}$    & $83.8_{0.4}$           & 10.2K/11.0K \\
Adapter (128)       & $66.1_{0.9}$    & $56.8_{0.6}$    & $83.3_{0.2}$           & 14.9K/15.6K \\
Adapter (64)        & $66.7_{0.5}$    & $56.2_{0.6}$    & $83.2_{0.1}$           & 24.1K/24.8K \\
HPT (4)             & $69.8_{0.6}$    & $52.8_{2.2}$    & $85.0_{0.1}$           & 7.9K/8.7K \\
HPT (8)             & $\mathbf{70.4_{0.6}}$ & $53.9_{1.4}$    & $86.4_{0.5}$      & 11.0K/11.8K \\
HPT (16)            & $70.0_{0.2}$    & $\mathbf{60.1_{3.1}}$ & $88.1_{0.2}$           & 17.2K/18.0K \\
\bottomrule
\end{tabular}
\end{table}

\begin{figure}[t]
    \centering
    \includegraphics[width=0.83\columnwidth]{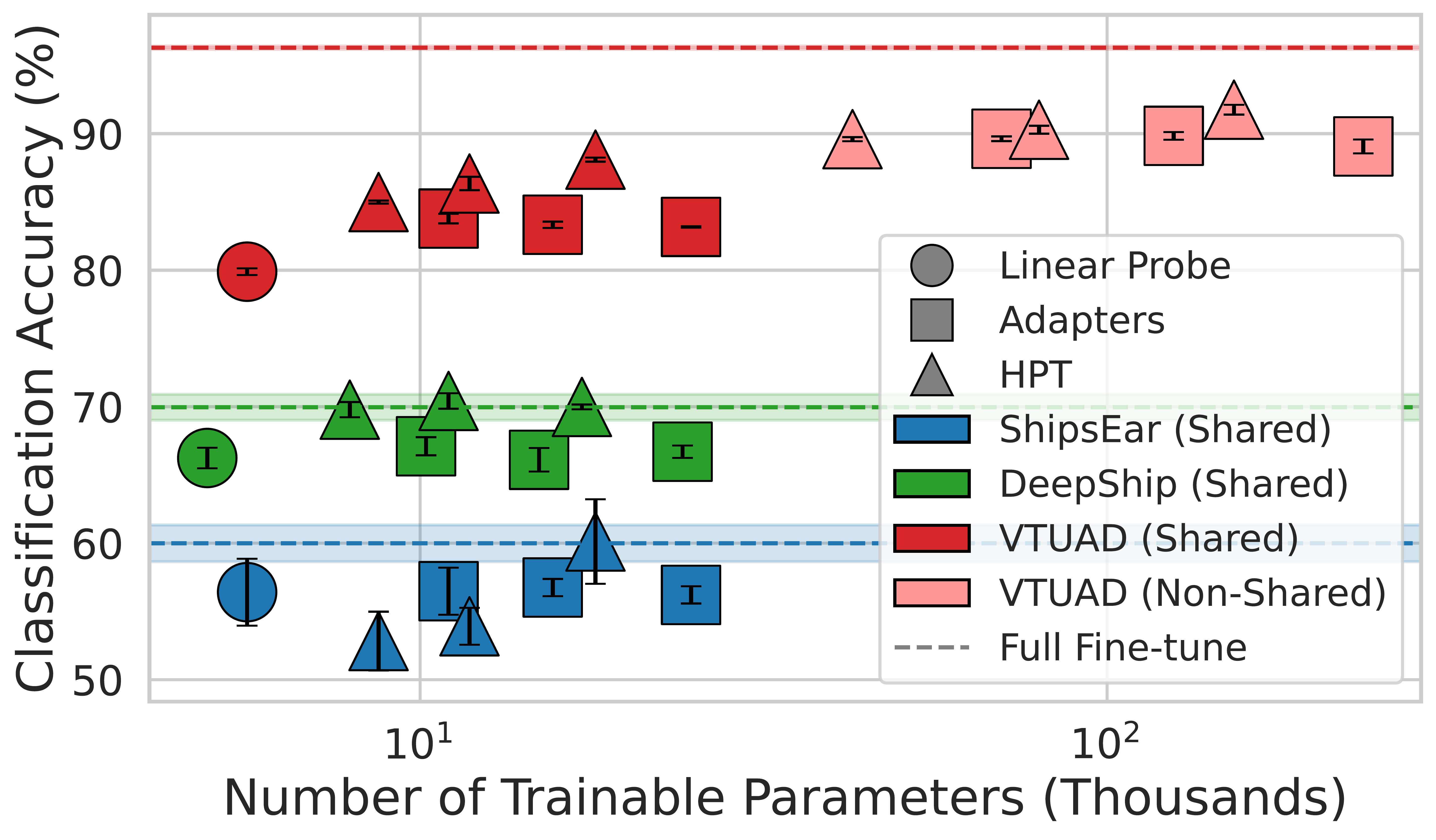}
    \caption{Scalability comparison of HPT and adapters. HPT performance increases with the number of bins if it has not already reached the full fine-tuning accuracy.}
    \label{fig:comparison}
\end{figure}

Since full fine-tuning accuracy was not achieved for VTUAD, the shared-weight restriction was removed, allowing each layer to have its own adapter or histogram parameters. Table \ref{tab:results_nonshared} shows that with non-shared weights, histogram configurations more closely approach full fine-tuning accuracy. Allowing each layer to fine-tune its own histogram parameters provides more control over layer-specific distributional alignment. Fig. \ref{fig:comparison} visualizes HPT's scalable performance with the number of trainable parameters, while performance of the adapters saturates. 

\begin{table}[ht]
\centering
\caption{Mean classification accuracy (\%) and parameters on VTUAD under non-shared weights settings. Standard deviations are subscripted. The best accuracy is shown in \textbf{bold}.}
\label{tab:results_nonshared}
\renewcommand{\arraystretch}{1.2}
\begin{tabular}{l c c}
\toprule
Method            & VTUAD Accuracy                    & Params \\
\midrule
Full fine tune  & $\mathbf{96.3_{0.1}}$        & 86.9M        \\
Linear probe    & $79.9_{0.2}$                 & 5.6K         \\
Adapter (256)     & $89.6_{0.2}$                 & 70.2K        \\
Adapter (128)     & $89.8_{0.3}$                 & 125.0K       \\
Adapter (64)      & $89.1_{0.5}$                 & 236.0K       \\
HPT (4)           & $89.6_{0.2}$                 & 42.6K        \\
HPT (8)           & $90.3_{0.3}$                 & 79.6K        \\
HPT (16)          & $91.8_{0.4}$                 & 153.0K       \\
\bottomrule
\end{tabular}
\end{table}

\begin{figure}[t]
    \centering
    \includegraphics[width=.81\columnwidth]{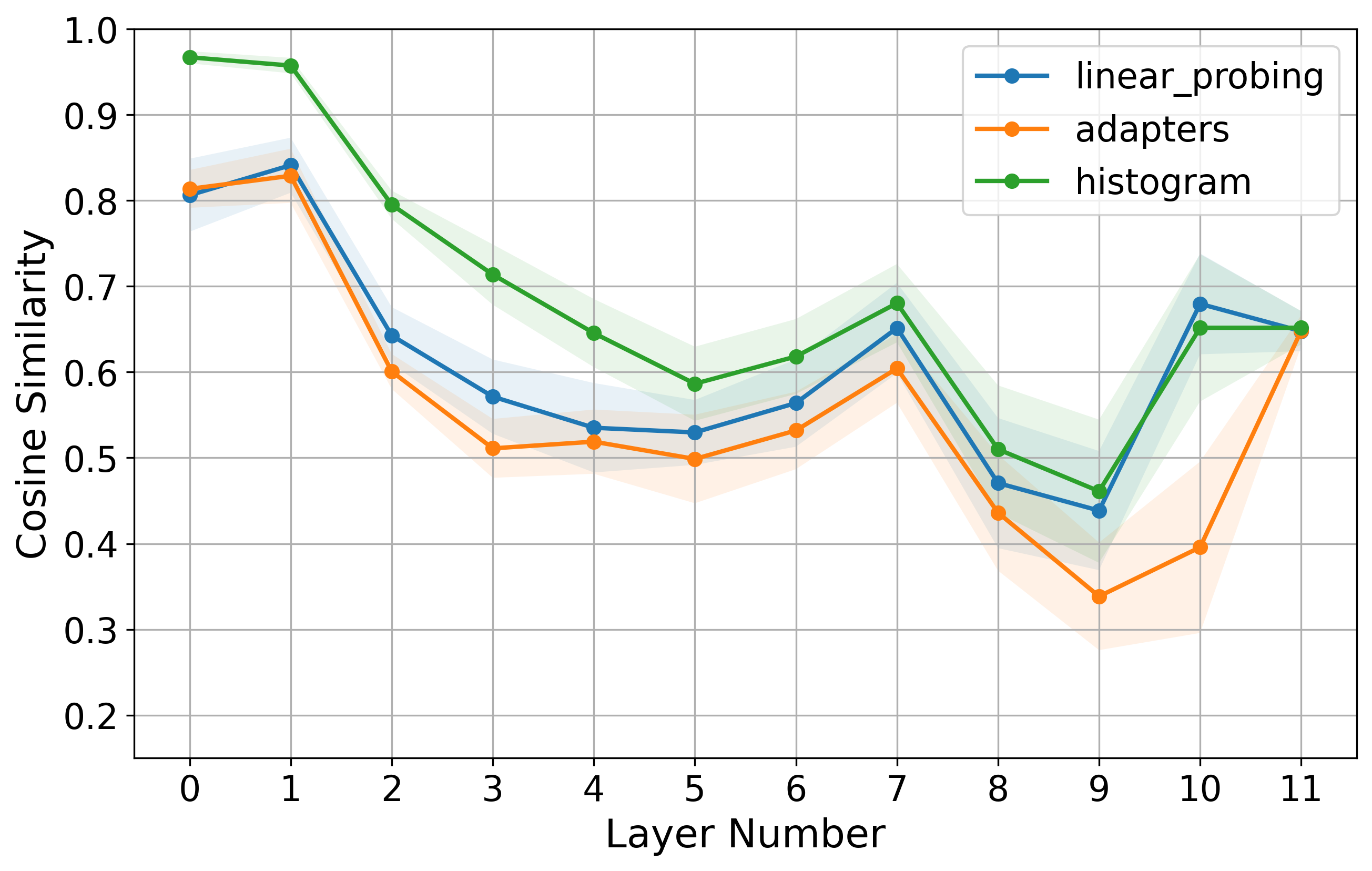}
    \caption{Layer‑wise feature similarity for VTUAD (non-shared). HPT maintains closer similarity with the fully fine‑tuned model’s representations.}
    \label{fig:feature_similarity}
\end{figure}

Further analysis is conducted with the largest dataset (VTUAD) using non-shared weights. Fig. \ref{fig:feature_similarity} visualizes feature similarity analysis \cite{lian2022scaling}, which compares the layer-wise similarity of linear probe, adapter and histogram methods to the fully fine-tuned baseline. It is observed that HPT retains higher similarity, indicating that it better preserves semantic representations than adapters, while adapters sometimes even achieve a lower similarity than linear probing. It is observed that in the early layers, all methods preserve low-level features, thereby maintaining high similarity. Progressing into the mid-layers, where more specialized abstractions occur, the adapter approach deviates more strongly, causing similarity to drop. In contrast, histogram-based methods, by summarizing the distribution, retain closer alignment with the baseline’s intermediate features. Finally, at the upper layers, both methods converge somewhat again, as the model refines representations for the final classification task.

To compare HPT with other related works, LoRA \cite{hu2021lora} and SSF \cite{lian2022scaling} were integrated into AST. For LoRA, the mechanism is applied to the query component with a rank of 6 and 12, while \(\alpha\) is set to 1 so the scaling factor becomes \(1/6\) and \(1/12\), respectively. For SSF, six layers were inserted at these points: after the first LayerNorm, after the QKV projection, after the second LayerNorm, after the first and second fully connected layers. An element-wise affine transformation \( x \leftarrow \mathrm{scale} \odot x + \mathrm{shift} \) is learned. Shared settings across all experiments are used so that the performance differences arise exclusively from the fine-tuning mechanism itself with minimal parameter footprint. It should be noted that the SSF six layers are distinct in the block. 

\begin{table}[t]
\centering
\caption{Accuracy (\%) and trainable parameters for different shared fine-tuning methods. Standard deviations are subscripted. The best accuracies are shown in \textbf{bold}. In the Params column, the right number corresponds to ShipsEar and VTUAD and the left number corresponds to DeepShip.}
\label{tab:comp-shared-results-singlecol}
\renewcommand{\arraystretch}{1.2}
\begin{tabular}{l c c c c}
\toprule
\textbf{Method} & \textbf{DeepShip} & \textbf{ShipsEar} & \textbf{VTUAD} & \textbf{Params} \\
\midrule
\textbf{Full fine tune} & 70.0\textsubscript{1.0} & 60.0\textsubscript{1.4} &  $\mathbf{96.3_{0.1}}$ & 86.9M/86.9M \\
\textbf{Linear probe}    & 66.2\textsubscript{0.8} & 56.4\textsubscript{2.5} & 79.9\textsubscript{0.2} & 4.9K/5.6K  \\
\midrule
\textbf{LoRA} \\
rank of 6  & 68.0\textsubscript{1.5} & 45.3\textsubscript{2.1} & 82.6\textsubscript{0.8} & 14.1K/14.9K \\
rank of 12 & 67.8\textsubscript{0.5} & 47.9\textsubscript{1.1} & 84.3\textsubscript{0.6} & 23.3K/24.1K \\
\midrule
\textbf{SSF} \\
LayerNorm        & 66.2\textsubscript{0.2} & 56.9\textsubscript{0.3} & 86.3\textsubscript{0.3} & 6.4K/7.2K  \\
MHSA                & 67.9\textsubscript{0.7} & 55.4\textsubscript{1.2} & 90.5\textsubscript{0.5} & 12.5K/13.3K \\
MHSA \& FFN            & 68.4\textsubscript{0.7} & 54.8\textsubscript{1.1} & 92.2\textsubscript{0.3} & 21.8K/22.5K \\
\midrule
\textbf{HPT (ours)} \\
4 bins MHSA           & 69.8\textsubscript{0.6} & 52.8\textsubscript{2.2} & 85.0\textsubscript{0.1} & 7.9K/8.7K  \\
8 bins MHSA          & $\mathbf{70.4_{0.6}}$ & 53.9\textsubscript{1.4} & 86.4\textsubscript{0.5} & 11.0K/11.8K \\
16 bins MHSA           & 70.0\textsubscript{0.2} & $\mathbf{60.1_{3.1}}$ & 88.1\textsubscript{0.2} & 17.2K/18.0K \\
\bottomrule
\end{tabular}
\end{table}

The results in Table \ref{tab:comp-shared-results-singlecol} show varying strengths across methods and datasets. On DeepShip, HPT achieves the highest accuracy (70.4\%). This indicates that distribution‐based tuning can effectively capture the acoustic characteristics of mid‐scale datasets. In contrast, on VTUAD, the larger dataset size favors SSF with six layers (92.2\%). This reinforces the idea that no single tuning approach dominates.

The results in Table \ref{tab:sonar-peft-results} show that generally most of the methods reach high accuracy on active sonar datasets. On Watertank, SSF with three insertion points at MHSA attains the best mean (98.99). HPT with 8 bins is broadly comparable to SSF. On Turntable, SSF with 6 insertion points matches full fine-tuning. Given the controlled data acquisition protocols, the frozen model features already linearly separate most classes (Watertank linear probe at 97.6), so PETL methods require smaller corrections. HPT with no insertion-site choices (unlike SSF) reaches competitive values, whereas the LoRA setup here appears too restrictive for Turntable’s viewpoint variation.

\begin{table}[ht]
\centering
\caption{Active sonar: Accuracy (\%) and trainable parameters on Watertank and Turntable. Standard deviations are subscripted. Best accuracies per dataset are in \textbf{bold}. In the Params column, the left number corresponds to Watertank and the right number to Turntable.}
\label{tab:sonar-peft-results}
\renewcommand{\arraystretch}{1.2}
\begin{tabular}{l c c c}
\toprule
\textbf{Method} & \textbf{Watertank} & \textbf{Turntable} & \textbf{Params} \\
\midrule
\textbf{Full fine tune} & 98.57\textsubscript{0.24} & \textbf{97.47\textsubscript{0.70}} & 86.9M/86.9M \\
\textbf{Linear probe}   & 97.64\textsubscript{0.52} & 92.61\textsubscript{0.19} & 10.0K/10.8K \\
\midrule
\multicolumn{4}{l}{\textbf{LoRA}}\\
rank 6  & 97.38\textsubscript{0.24} & 88.22\textsubscript{0.51} & 19.2K/20.0K \\
rank 12 & 96.37\textsubscript{0.63} & 86.25\textsubscript{0.19} & 28.4K/29.2K \\
\midrule
\multicolumn{4}{l}{\textbf{SSF}}\\
LayerNorm & 98.40\textsubscript{0.32} & 95.81\textsubscript{0.25} & 11.5K/12.3K \\
MHSA & \textbf{98.99\textsubscript{0.21}} & 97.26\textsubscript{0.19} & 17.7K/18.4K \\
MHSA \& FFN & 98.48\textsubscript{0.21} & \textbf{97.47\textsubscript{0.96}} & 26.9K/27.7K \\
\midrule
\multicolumn{4}{l}{\textbf{HPT (ours)}}\\
4 bins MHSA & 98.23\textsubscript{0.21} & 94.94\textsubscript{0.51} & 13.1K/13.8K \\
8 bins MHSA & 98.40\textsubscript{0.43} & 96.02\textsubscript{0.74} & 16.2K/16.9K \\
16 bins MHSA & 98.06\textsubscript{0.52} & 95.87\textsubscript{0.62} & 22.3K/23.1K \\
\bottomrule
\end{tabular}
\end{table}

\section{Conclusion}
\label{conc}

In this work, a novel parameter-efficient tuning method, a histogram layer integrated into the transformer architecture, is introduced. Experiments on three passive sonar datasets demonstrated that HPT not only achieves competitive accuracy compared to conventional adapters, but does so with a favorable parameter trade-off. 
Extending to active sonar, HPT is broadly competitive to other methods, while SSF with multiple insertion points attains the best results. There remain some avenues for further research. HPT's efficacy can depend on hyper-parameters such as the number of bins. Future work could explore adaptive binning algorithms. Another promising direction lies in combining HPT with other PETL techniques (e.g., LoRA), by joint optimization to leverage complementary benefits. This work aims to pave the way for more distribution-aware techniques in efficient learning strategies.  

\section*{Acknowledgment}
Portions of this research were conducted with the advanced computing resources provided by Texas A\&M High Performance Research Computing. 

\balance
\small
\bibliographystyle{IEEEtranN}
\bibliography{references}

\end{document}